\documentclass[runningheads]{llncs}
\usepackage{graphicx}

\usepackage{tikz}
\usepackage{comment}
\usepackage{amsmath,amssymb} 
\usepackage{color}
\usepackage{hyperref}
\usepackage{multirow}
\usepackage[capitalize]{cleveref}
\usepackage{bbm}
\usepackage{bm}

\usepackage[accsupp]{axessibility}  

\usepackage{wrapfig}
\usepackage{xspace}
\newcommand{\mypara}[1]{\noindent{\bf{#1}}}
\newcommand{\modelName}{\textsc{TCaF}\xspace}
\newcommand{\activity}{{ActivityNet-GZSL$^{cls}$}\xspace}
\newcommand{\ucf}{{UCF-GZSL$^{cls}$}\xspace}
\newcommand{\vgg}{{VGGSound-GZSL$^{cls}$}\xspace}
\newcommand{\selaactivity}{{ActivityNet-GZSL}\xspace}
\newcommand{\selaucf}{{UCF-GZSL}\xspace}
\newcommand{\selavgg}{{VGGSound-GZSL}\xspace}

\usepackage{lipsum}

\makeatletter
\newcommand{\printfnsymbol}[1]{%
  \textsuperscript{\@fnsymbol{#1}}%
}
\makeatother

\begin{document}

\pagestyle{headings}
\mainmatter
\def\ECCVSubNumber{6942}  

\title{Temporal and cross-modal attention for audio-visual zero-shot learning} 

\titlerunning{Temporal and cross-modal attention for audio-visual ZSL}
%
\newcommand*\samethanks[1][\value{footnote}]{\footnotemark[#1]}
\author{Otniel-Bogdan Mercea\thanks{Denotes equal contribution}\inst{1}\index{Mercea, Otniel-Bogdan} \and
Thomas Hummel\samethanks\inst{1} \and
\mbox{A. Sophia Koepke}\inst{1}\index{Koepke, A. Sophia} \and
Zeynep Akata\inst{1,2,3}}
\authorrunning{O.-B. Mercea et al.}
%
\institute{University of T{\"u}bingen \and
MPI for Informatics \and
MPI for Intelligent Systems\\
\email{\{otniel-bogdan.mercea, thomas.hummel, a-sophia.koepke,zeynep.akata\}@uni-tuebingen.de}}

\maketitle

\begin{abstract}
Audio-visual generalised zero-shot learning for video classification requires understanding the relations between
the audio and visual information in order to be able to recognise samples from novel, previously unseen classes at test time. 
The natural semantic and temporal alignment between audio and visual data in video data can be exploited to learn powerful representations that generalise to unseen classes at test time.
We propose a multi-modal and Temporal Cross-attention Framework (\modelName) for audio-visual generalised zero-shot learning. Its inputs are temporally aligned audio and visual features that are obtained from pre-trained networks. Encouraging the framework to focus on cross-modal correspondence across time instead of self-attention within the modalities boosts the performance significantly. We show that our proposed framework that ingests temporal features yields state-of-the-art performance on the \ucf, \vgg, and \activity benchmarks for (generalised) zero-shot learning. Code for reproducing all results is available at \url{https://github.com/ExplainableML/TCAF-GZSL}.
\keywords{Zero-shot learning, Audio-visual learning}
\end{abstract}

\section{Introduction}
Learning task-specific audio-visual representations commonly requires a great number of annotated data samples.
However, annotated datasets are limited in size and in the labelled classes that they contain. If a model which was trained with supervision on such a dataset is applied in the real world, it encounters classes that it has never seen. To recognise those novel classes, it would not be feasible to train a new model from scratch. Therefore, it is essential to analyse the behaviour of a trained model in new settings. Ideally, a model should be able to transfer knowledge obtained from classes seen during training to previously unseen categories. This ability is probed in the zero-shot learning (ZSL) task. In addition to zero-shot capabilities, a model should retain the class-specific information from seen training classes. This is challenging and is investigated in the so-called generalised ZSL (GZSL) setting which considers the performance on both, seen and unseen classes. 

Prior works~\cite{parida2020coordinated,mazumder2021avgzslnet,mercea2022} have proposed frameworks that address the (G)ZSL task for video classification using audio-visual inputs. Those methods learn a mapping from the audio-visual input data to textual label embeddings, enabling the classification of samples from unseen classes. At test time, the class whose word embedding is closest to the predicted audio-visual output embedding is selected. Similar to this, we use the textual label embedding space to allow for information transfer from training classes to previously unseen classes.
\begin{wrapfigure}{r}{0.65\linewidth}
\centering
   \includegraphics[clip, trim=30 10 30 5 ,width=\linewidth]{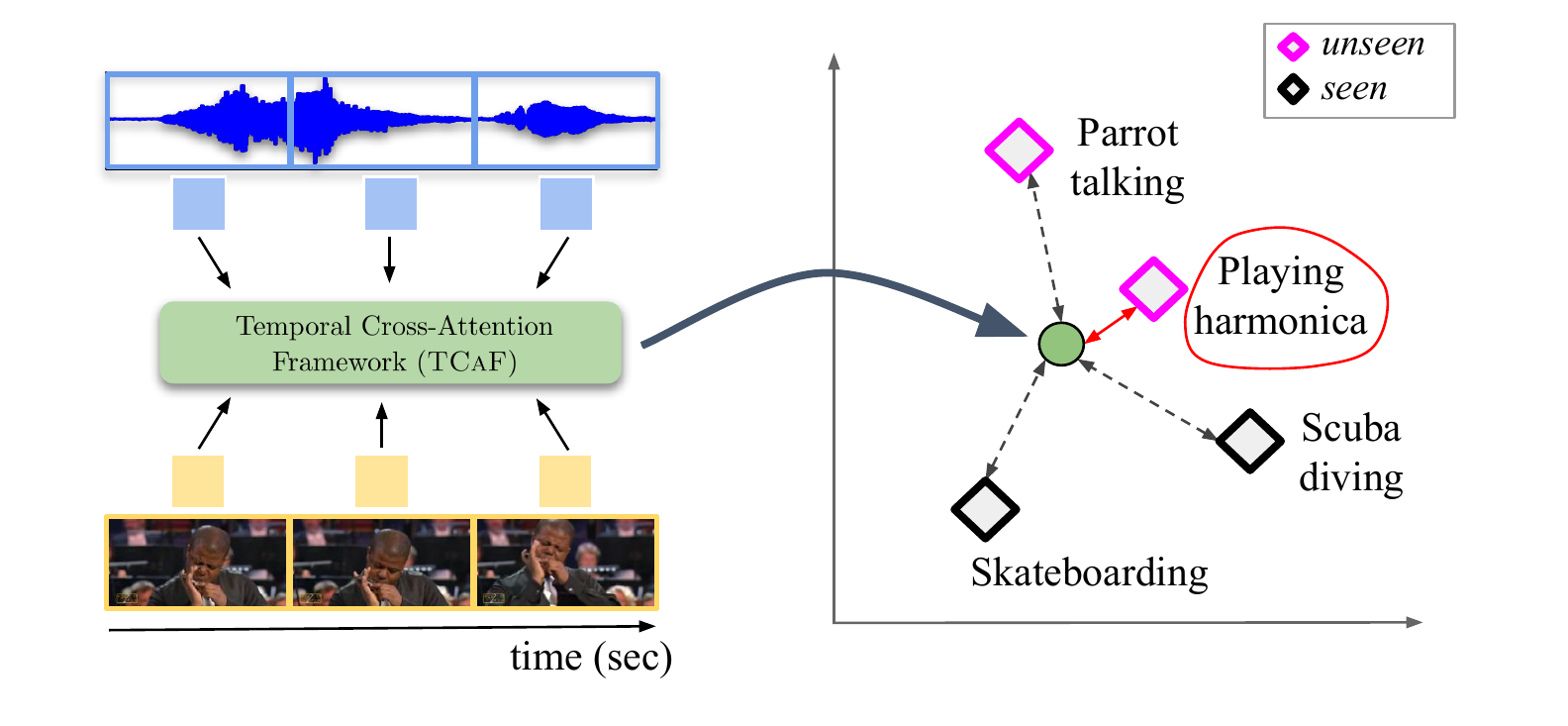}
    \caption{Our temporal cross-attention framework for audio-visual (G)ZSL learns a multi-modal embedding (green circle) by exploiting the temporal alignment between audio and visual data in videos. Textual label embeddings (grey squares) are used to transfer information from seen training classes (black) to unseen test classes (pink). The correct class is playing harmonica (red).}
    \label{fig:teaser}
\end{wrapfigure}
 However, \cite{parida2020coordinated,mazumder2021avgzslnet,mercea2022} used temporally averaged features as inputs that were extracted from networks pre-trained on video data. The averaging disregarded the temporal dynamics in videos.
We propose a Temporal Cross-attention Framework (\modelName) which builds on \cite{mercea2022} and additionally exploits temporal information by using temporal audio and visual data as inputs. This gives a significant boost in performance for the audio-visual (G)ZSL task compared to using temporally averaged input features. Different from computationally expensive methods that operate directly on raw visual inputs~\cite{brattoli2020rethinking,lin2022cross,kerrigan2021reformulating}, our \modelName uses features extracted from networks pre-trained for audio and video classification as inputs. This leads to an efficient setup that uses temporal information instead of averaging across time.

The natural alignment between audio and visual information in videos, e.g.\
a frog being visible in a frame while the sound of a frog croaking is audible, provides a rich training signal for learning video representations. This can be attributed to the semantic and temporal correlation between the audio and visual information when comparing the two modalities. We encourage our \modelName to put special emphasis on the correlation across the two modalities by employing repeated cross-attention. This attention mechanism only allows attention to tokens from the other modality. This effectively acts as a bottleneck which results in cheaper computations and gives a boost in performance over using full self-attention across all tokens from both modalities.

We perform a detailed model ablation study to show the benefits of using temporal inputs and our proposed cross-attention. Furthermore, we confirm that our training objective is well-suited to the task at hand. We also analyse the learnt audio-visual embeddings with t-SNE visualisations which confirm that training our \modelName improves the class separation for both seen and unseen classes.

To summarise, our contributions are as follows: (1) We propose a temporal cross-attention framework \modelName for audio-visual (G)ZSL. (2) Our proposed model achieves state-of-the-art results on the \ucf, \vgg, and \activity datasets, demonstrating that using temporal information is extremely beneficial for improving the (generalised) zero-shot classification accuracy compared to using temporally averaged features as model inputs. (3) We perform a detailed analysis of the use of enhanced cross-attention across modalities and time, demonstrating the benefits of our proposed model architecture and training setup. 

\section{Related work}
Our work relates to several themes in the literature: audio-visual learning, ZSL with side information, audio-visual ZSL with side information, and multi-modal transformer architectures. We discuss those in more detail in the following.

\mypara{Audio-visual learning.}
The temporal alignment between audio and visual data in videos is a strong learning signal which can be exploited for learning audio-visual representations.~\cite{owens2016ambient,owens2018learning,alwassel2019self,patrick2020multi,korbar2018cooperative,aytar2016soundnet}. In addition to audio and video classification, numerous other tasks benefit from audio-visual inputs, such as the separation and localisation of sounds in video data~\cite{owens2018audio,tian2018audio,arandjelovic2018objects,gao2019co,chen2021localizing,Afouras20b,afouras2021selfsupervised}, audio-driven synthesis of images \cite{wiles2018x2face,jamaludin2019you}, audio synthesis driven by visual information \cite{zhou2019vision,goldstein2018guitar,koepke2019visual,koepke2020sight,su2020multi,gan2020foley,narasimhan2021strumming}, and lip reading~\cite{afouras2020asr,afouras2018deep}. Some approaches use class-label supervision between modalities \cite{fayek2020large,chen2021distilling} which does not require the temporal alignment between the input modalities. In contrast to full class-label supervision, we train our model only on the subset of seen training classes. 

\mypara{ZSL with side information.} Visual ZSL methods commonly map the visual inputs to class side information~\cite{frome2013devise,akata2015evaluation,akata2015label}, e.g.\ word2vec~\cite{mikolov2013efficient} class label embeddings. This allows to determine the class with the side information that is closest at test time as the class prediction. Furthermore, attribute annotations have been used as side information~\cite{wah2011caltech,xiao2010sun,xian2018zero,farhadi2009describing}. 
Recent non-generative methods identify key visual attributes \cite{xu2020attribute}, use attention to find discriminative regions \cite{xie2019attentive}, or disambiguate class embeddings \cite{liu2019attribute}. In contrast, feature generation methods 
train a classifier on generated and real features \cite{xian2019f,narayan2020latent,zhu2019learning,xian2018feature}. Unlike methods for ZSL with side information with unimodal (visual) inputs, our proposed framework uses multi-modal audio-visual inputs.

\mypara{Audio-visual ZSL with side information.} The task of GZSL from audio-visual data was introduced by~\cite{parida2020coordinated,mazumder2021avgzslnet} on the AudioSetZSL dataset~\cite{parida2020coordinated} using class label word embeddings as side information. Recently, \cite{mercea2022} proposed the AVCA framework which uses cross-attention to fuse information from the averaged audio and visual input features for audio-visual GZSL. Our proposed framework builds on \cite{mercea2022}, but instead of using temporally averaged features as inputs~\cite{mercea2022,parida2020coordinated,mazumder2021avgzslnet}, we explore the benefits of using temporal cross-attention information. Unlike \cite{mercea2022}'s two-stream architecture, we propose the fusion into a single output branch with a classification token that aggregates multi-modal information. Furthermore, we simplify the training objective, and show that the combination of using temporal inputs, our architecture, and training setup leads to superior zero-shot classification performance.

\mypara{Multi-modal transformers.}
The success of transformer models in the language domain~\cite{vaswani2017attention,devlin2018bert,radford2019language} has been translated to visual recognition tasks with the Vision Transformer~\cite{dosovitskiy2020image}. 
Multi-modal vision-language representations have been obtained with a masked language modelling objective, and achieved state-of-the-art performance on several text-vision tasks~\cite{sun2019learning,sun2019videobert,lu2019vilbert,li2020unicoder,li2019visualbert,su2019vl,tan2019lxmert}. In this work, we consider audio-visual multi-modality.
Transformer-based models that operate on audio and visual inputs have recently been proposed for text-based video retrieval~\cite{gabeur2020multi,liu2021hit,wang2021t2vlad}, dense video captioning \cite{iashin2020better}, audio-visual event localization \cite{lin2020audiovisual}, and audio classification \cite{boes2019audiovisual}. Different to vanilla transformer-based attention, our \modelName puts special emphasis on cross-attention between the audio and visual modalities in order to learn powerful representations for the (G)ZSL task.

\section{\modelName Model}
In this section, we describe the problem setting (\cref{sec:problem_setting}), our proposed model architecture (\cref{sec:architecture}), and the loss functions used to train \modelName (\cref{sec:loss_functions}).

\begin{figure}[t]
    \centering
    \includegraphics[width=0.99\linewidth]{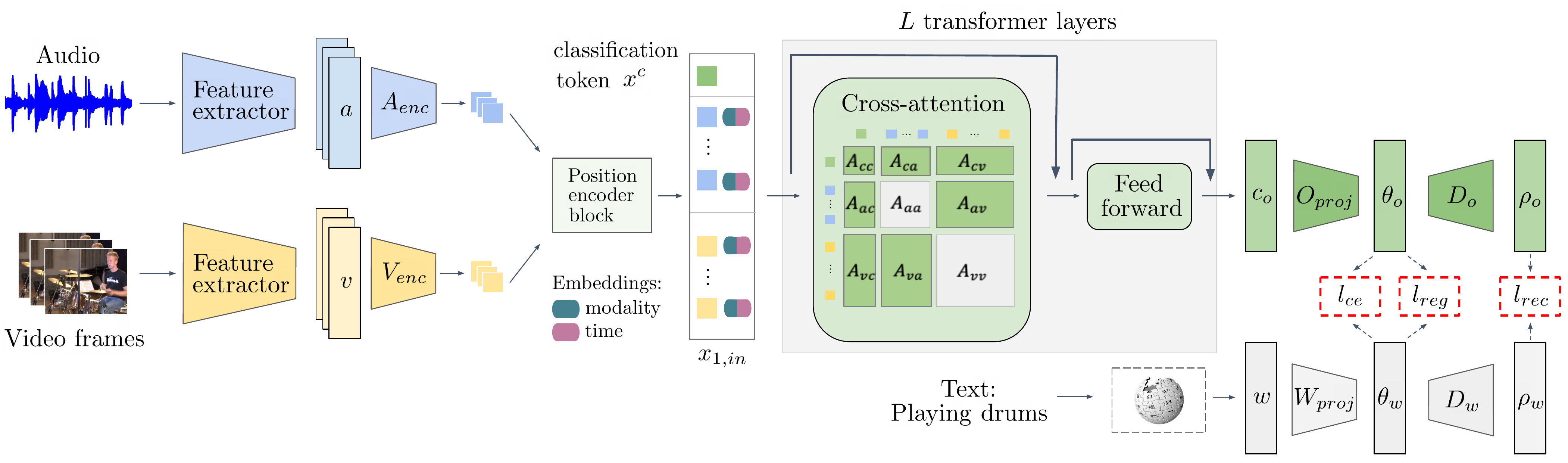}
    \caption{\modelName takes audio and visual features extracted from video data as inputs. Those are embedded and equipped with modality and time embeddings before passing through a sequence of $L$ transformer layers with cross-attention. The output classification token $c_o$ is then projected to embedding spaces that are shared with the textual information. The loss functions operate on the joint embedding spaces. At test time, the class prediction $c$ is obtained by determining the word label embedding $\theta_w^j$ that is closest to $\theta_o$.
    }
    \label{fig:architecture}
\end{figure}

\subsection{Problem setting}\label{sec:problem_setting}
We address the task of (G)ZSL using audio-visual inputs. The aim of ZSL is to be able to generalise to previously unseen test classes at test time. For GZSL, the model should additionally preserve knowledge about seen training classes, since the GZSL test set contains samples from both, seen and unseen classes.

We denote an audio-visual dataset with $N$ samples and $K$ (seen and unseen) classes by $\mathcal{V} = \{\mathcal{X}_{\bm{a}[i]}, \mathcal{X}_{\bm{v}[i]}, y_{[i]}\}_{i=1}^{N}$, consisting of audio data $\mathcal{X}_{\bm{a}[i]}$, visual data $\mathcal{X}_{\bm{v}[i]}$, and ground-truth class labels $y_{[i]} \in \mathbb{R}^{{K}}$. Naturally, video data contains temporal information. In the following, we use $T_a$ and $T_v$ to denote the number of audio and visual segments in a video clip.

A pre-trained audio classification CNN is used to extract a sequence of audio features $\bm{a}_{[i]} = \{a_{1}, \dots, a_{t}, \dots, a_{T_a}\}_i$ to encode the audio information $\mathcal{X}_{\bm{a}[i]}$.
The visual data $\mathcal{X}_{\bm{v}[i]}$ is encoded into a temporal sequence of features $\bm{v}_{[i]} = \{v_{1}, ..., v_{t}, ..., v_{T_v}\}_i$ by representing visual segments with features extracted from a pre-trained video classification network.

\subsection{Model architecture}\label{sec:architecture}
In the following, we describe the architecture of our proposed \modelName (see \cref{fig:architecture}). 

\mypara{Embedding the inputs and position encoder block.}
\modelName takes pre-extracted audio and visual features $\bm{a}_{[i]}$ and $\bm{v}_{[i]}$ as inputs. For readability, we will drop the subscript $i$ in the following which denotes the $i-$th sample.
In order to project audio and visual features to the same feature dimension, $\bm{a}$ and $\bm{v}$ are passed through two modality-specific embedding blocks, giving embeddings:
\begin{equation}
    \phi_a = A_{enc}(\bm{a})  \text{ and } \phi_v = V_{enc}(\bm{v}),
\end{equation} 
with $\phi_a \in \mathbb{R}^{T_a * d_{dim}}$ and $\phi_v \in \mathbb{R}^{T_v * d_{dim}}$ . The embedding blocks are composed of two linear layers $f^m_1, f^m_2$ for $m \in \{\bm{a},\bm{v}\}$, where $f^m_1:\mathbb{R}^{T_m*d_{in_m}} \rightarrow \mathbb{R}^{T_m*d_{fhidd}}$ and $f^m_2:\mathbb{R}^{T_m*d_{fhidd}} \rightarrow \mathbb{R}^{T_m*d_{dim}}$. $f^m_1, f^m_2$ are each followed by batch normalisation~\cite{ioffe2015batch}, a ReLU~\cite{nair2010rectified}, and dropout~\cite{JMLR:v15:srivastava14a} with dropout rate $drop_{enc}$.

The position encoder block adds learnt modality and temporal positional embeddings to the outputs of the modality-specific embedding blocks. We explain this in detail below.
To handle different frame rates in the audio and visual modalities, we use Fourier features~\cite{tancik2020fourier} $pos_t \in \mathbb{R}^{d_{pos}}$ for the temporal embeddings that encode the actual point in time in the video which corresponds to an audio or visual representation. This allows to capture the relative temporal position of the audio and visual features across the modalities.

For an audio embedding $\phi_{a_t}$ at time $t$, a linear map $g_a: \mathbb{R}^{d_{pos} + d_{dim}} \rightarrow \mathbb{R}^{d_{dim}}$, and a dropout layer $g^{D}$ with dropout probability $drop_{prob,pos}$, we obtain position-aware audio feature tokens
\begin{equation}
    a^p_t = g^D(g_a(concat(\phi_{a_t},pos_{at}))) \qquad \text{with} \qquad pos_{at} = pos_a + pos_t,
\end{equation}
with modality and temporal embeddings $pos_a, pos_t \in \mathbb{R}^{d_{pos}}$ respectively. Position-aware visual tokens $v^p_t$ are obtained analogously.

Furthermore, we prepend a learnt classification token $x^c \in \mathbb{R}^{d_{dim}}$ to the sequence of feature tokens. The corresponding output classification token $c_o$ is used by our output projection $O_{proj}$ to obtain the final prediction.

\mypara{Audio-visual transformer layers.}
\modelName contains $L$ stacked audio-visual transformer layers that allow for enhanced cross-attention. Each of our transformer layers consists of an attention function $f_{l,Att}$, followed by a feed forward function $g_{l,FF}$. The output of the $l$-th transformer layer is given as
\begin{equation}
    x_{l,out} = x_{l,ff} + x_{l,att} = g_{l,FF}(x_{l,att}) + x_{l,att}, 
\end{equation}
with 
\begin{equation}
    x_{l,att} = f_{l,Att}(x_{l,in})+x_{l,in},
\end{equation}
where
\[
    x_{l,in}= 
\begin{cases}
    [x^c,a^p_1,\cdots,a^p_{T_a},v^p_1,\cdots,v^p_{T_v}] & \text{if } l = 1, \\
    x_{l-1,out} & \text{if } 2 \geq l \leq L.\\
\end{cases}
\]
We explain the cross-attention used in our transformer layers in the following.

\mypara{Transformer cross-attention.} \modelName{} primarily exploits cross-modal audio-visual attention to combine the information across the audio and visual modalities. All attention mechanisms in \modelName{} consist of multi-head attention~\cite{vaswani2017attention} with $H$ heads and a dimension of $d_{head}$ per head. 

We describe the first transformer layer $\mathcal{M}_1$, the transformer layer $\mathcal{M}_l$ operates analogously. 
We project the position-aware input features $x^c$, $\{a_t^p\}_{t \in [1,T_a]}$, $\{v_t^p\}_{t \in [1,T_v]}$ to queries, keys, and values with linear maps $g_s: \mathbb{R}^{d_{dim}} \xrightarrow{} \mathbb{R}^{d_{head} H}$ for $s \in \{q,k,v\}$. We can then write the outputs of the projection as zero-padded query, key, and value features. We write those out for the queries below, the keys and values are padded in the same way:
\begin{align}
    \mathbf{q}_c &= [g_q(x^c), 0, \cdots, 0],\\
    \mathbf{q}_a &= [0,\cdots,0,g_q(a^p_1),\cdots,g_q(a^p_{T_a}), 0, \cdots, 0], \\
     \mathbf{q}_v &= [0,\cdots,0,g_q(v^p_1),\cdots,g_q(v^p_{T_v})].
\end{align}
The full query, key, and value representations, $\mathbf{q}$, $\mathbf{k}$, and $\mathbf{v}$, are the sums of their modality-specific components
\begin{align}\label{eq:modality_components}
    \mathbf{q} = \mathbf{q}_c + \mathbf{q}_a + \mathbf{q}_v, \qquad 
    \mathbf{k} = \mathbf{k}_c + \mathbf{k}_a + \mathbf{k}_v, \qquad \text{and   } 
    \mathbf{v} = \mathbf{v}_c + \mathbf{v}_a + \mathbf{v}_v.
\end{align} 
The output of the first attention block $x_{1,att}$ is the aggregation of the per-head attention with a linear mapping $g_h: \mathbb{R}^{d_{head} H} \rightarrow \mathbb{R}^{d_{dim}}$, $g^{DL}$ dropout with dropout probability $drop_{prob}$ and layer normalisation $g^{LN}$ \cite{ba2016layer}, such that
\begin{equation}
    x_{1,att} = f_{1,Att}(x_{l,in}) = g^{DL}(g_h(f_{1,att}^1(g^{LN}(x_{1,in})),\cdots,f_{1,att}^H(g^{LN}(x_{1,in})))),
\end{equation}
with the attention $f^h_{att}$ for the attention head $h$.
We can write the attention for the head $h$ as 
\begin{equation}
    f^h_{att}(x_{1,in}) = softmax \left( \frac{\mathbf{A}}{\sqrt{d_{head}}} \right) \mathbf{v},
\end{equation}
where $\mathbf{A}$ can be split into its cross-attention and self-attention components:
\begin{equation}\label{eq:attention}
\mathbf{A}_c 
= \mathbf{q}_c \, \mathbf{k}^T + \mathbf{k} \, \mathbf{q}_c^T, \qquad 
\mathbf{A}_x =
\mathbf{q}_a \, \mathbf{k}_v^T + \mathbf{q}_v \, \mathbf{k}_a^T, \\
\end{equation}
\begin{equation*}
\mathbf{A}_{self} 
= \mathbf{q}_a \, \mathbf{k}_a^T + \mathbf{q}_v \, \mathbf{k}_v^T.
\end{equation*}
We then get
\begin{equation}
\mathbf{A} = \mathbf{A}_c + \mathbf{A}_x + \mathbf{A}_{self} 
= \begin{pmatrix}
A_{cc} & A_{ca} & A_{cv}\\
A_{ac} & \ddots & \vdots \\
A_{vc} & \hdots & 0
\end{pmatrix}
+ \begin{pmatrix}
0 & \hdots & 0 \\
\vdots & \ddots & A_{av} \\
0 & A_{va} & 0
\end{pmatrix}
+ \begin{pmatrix}
0 & \hdots & 0 \\
\vdots & A_{aa} & \vdots \\
0 & \hdots & A_{vv}
\end{pmatrix},
\end{equation}
where the $A_{mn}$ with $m,n \in \{c,a,v\}$ describe the attention contributions from the classification token, the audio and the visual modalities respectively.

Our \modelName uses the cross-attention $\mathbf{A}_c + \mathbf{A}_{x}$ to put special emphasis on the attention across modalities. 
Results for different model variants that use only the within-modality self-attention ($\mathbf{A}_c + \mathbf{A}_{self}$) or the full attention which combines self-attention and cross-attention are presented in \cref{sec:ablation}. 

\mypara{Feed forward function.} The feed forward function $g_{l,FF}: \mathbb{R}^{d_{dim}} \xrightarrow{} \mathbb{R}^{d_{dim}}$ is applied to the output of the attention function
\begin{equation}
    x_{l,ff} = g_{l, FF}(x_{l,att})=g^{DL}(g_{l,F2}(g^{DL}(g^{GD}(g_{l,F1}(g^{LN}(x_{l,att}))))))
\end{equation}
where $g_{l,F1}: \mathbb{R}^{d_{dim}} \xrightarrow{} \mathbb{R}^{d_{ff}}$ and $g_{l,F2}: \mathbb{R}^{d_{ff}} \xrightarrow{} \mathbb{R}^{d_{dim}}$ are linear mappings, $g^{GD}$ is a GELU layer~\cite{hendrycks2016gaussian} and a dropout layer with dropout probability $drop_{prob}$, $g^{DL}$ is dropout with $drop_{prob}$ and $g^{LN}$ is layer normalisation. 

\mypara{Output prediction.}
To determine the final class prediction, the audio-visual embedding is projected to the same embedding space as the textual class label representations.
We project the output classification token $c_o$ of the temporal cross-attention to $\theta_{o}=O_{proj}(c_o)$ where $\theta_{o} \in \mathbb{R}^{d_{out}}$. The projection block is composed of a sequence of two linear layers $f_3$ and $f_4$, where $f_3:\mathbb{R}^{d_{dim}} \rightarrow \mathbb{R}^{d_{fhidd}}$ and $f_4:\mathbb{R}^{d_{fhidd}} \rightarrow \mathbb{R}^{d_{out}}$. $f_3, f_4$ are each followed by batch normalisation, a ReLU, and dropout with rate $drop_{proj_o}$.
We project the word2vec class label embedding  $w^j$ for class $j$ using the projection block $W_{proj}(w^j)=\theta^j_w$, where $\theta^j_w \in \mathbb{R}^{d_{out}}$. $W_{proj}$ consists of a linear projection followed by batch normalisation, ReLU, and dropout with dropout rate $drop_{proj_w}$.
The class prediction $c$ is obtained by determining the projected word2vec embedding which is closest to the output embedding: 
\begin{equation}
    c = \underset{j}{\mathrm{argmin}}(\| \theta^j_w - \theta_{o} \|_2).
\end{equation}

\subsection{Loss functions}\label{sec:loss_functions}
Our training objective $l$ combines a cross-entropy loss $l_{ce}$, a reconstruction loss $l_{rec}$, and a regression loss $l_{reg}$:
\begin{equation}\label{eq:full_loss}
    l = l_{ce} + l_{rec} + l_{reg}.
\end{equation}

\mypara{Cross-entropy loss.}
For the ground-truth label $y_{i}$ with corresponding class index $k_{gt} \in \mathbb{R}^{K_{seen}}$,
the output of our temporal cross-attention $\theta_{{o}_i}$, and a matrix containing the textual label embeddings for the $K_{seen}$ seen classes $\theta_{w_{seen}}$, we define the cross-entropy loss for $n$ training samples as
\begin{equation}
l_{ce}=-\frac{1}{n} \sum_i^n y_i \log{ \left( \frac{\exp{(\theta_{w_{seen},{k_{gt}}} \theta_{o_i})}}{\sum_{k_j}^{K_{seen}} \exp{ (\theta_{w_{seen},{k_{j}}} \theta_{o_i}) }} \right)
}.
\end{equation}

\mypara{Regression loss.}
 While the cross-entropy loss updates the probabilities for both the correct and incorrect classes, our regression loss directly focuses on reducing the distance between the output embedding for a sample and the corresponding projected word2vec embedding. 
The regression loss is based on the mean squared error metric with the following formulation:
\begin{equation}\label{eq:reg_loss}
    l_{reg} = \frac{1}{n}\sum_{i=1}^n(\theta{o_{i}}-\theta_{w_{i}})^2,
\end{equation}
where $\theta_{o_{i}}$ is the audio-visual embedding, and $\theta_{w_{i}}$ is the projection of the word2vec embedding corresponding to the $i$-th sample.

\mypara{Reconstruction loss.}
The goal of the reconstruction loss is to ensure that the embeddings $\theta_{o}$ and $\theta_{w}$ contain semantic information from the word2vec embedding $w$. 
We use $D_{u}: \mathbb{R}^{d_{out}} \mapsto \mathbb{R}^{d_{dim}}$ with $\rho_{u}=D_{u}(\theta_u)$ for $u \in \{o,w\}$. $D_{w}$ is a sequence of one linear layer, batch normalisation, a ReLU, and dropout with rate $drop_{proj_w}$. $D_{o}$ is composed of a sequence of two linear layers each followed by batch normalisation, a ReLU, and dropout with dropout rate $drop_{proj_o}$.
Our reconstruction loss encourages the reconstruction of the output embedding, $\rho_{o_i}$, and the reconstruction of the word2vec projection, $\rho_{w_i}$, to be close to the original word2vec embedding $w_i$:
\begin{equation}\label{eq:pareto_mle2}
  l_{rec}=\frac{1}{n}\sum_{i=1}^n(\rho_{o_i}- w_i)^2+
  \frac{1}{n}\sum_{i=1}^n(\rho_{w_i}- w_i)^2.
\end{equation}

\section{Experiments}
In this section, we detail our experimental setup (\cref{sec:setup}), and compare to state-of-the-art methods for audio-visual GZSL (\cref{sec:sota}). Furthermore, we present an ablation study in~\cref{sec:ablation} which shows the benefits of using our proposed attention scheme and training objective. Finally, we present t-SNE visualisations of our learnt audio-visual embeddings in~\cref{sec:qualitative}.

\subsection{Experimental setup}\label{sec:setup}
Here, we describe the datasets used, the evaluation metrics, and the implementation details for all models.

\mypara{Datasets.}
We use the \ucf, \vgg, and \activity datasets~\cite{mercea2022} for audio-visual (G)ZSL for training and testing all models. \cite{mercea2022} introduced benchmarks for two sets of features, the first uses a model pre-trained using self-supervision on the VGGSound dataset from~\cite{asano2020labelling}, the second takes features extracted from pre-trained VGGish~\cite{hershey2017cnn} and C3D~\cite{tran2015learning} audio and video classification networks. Since the VGGSound dataset is also used for the zero-shot learning task (\selavgg), we selected the second option (using VGGish and C3D) and use the corresponding dataset splits proposed in~\cite{mercea2022}. We additionally provide results on the \selaucf, \selavgg, and \selaactivity datasets in the supplementary material.

In particular, the audio features are extracted using VGGish~\cite{hershey2017cnn} to obtain one 128-dimensional feature vector for each $0.96$\,s snippet. The 
visual features are obtained using C3D~\cite{tran2015learning} pre-trained on Sports-1M~\cite{KarpathyCVPR14}.
For this, all videos are resampled to 25\,fps. A 4096-dimensional feature vector is then extracted for 16 consecutive video frames.

\mypara{Evaluation metrics.}
We follow~\cite{xian2018zero,mercea2022} and use the mean class accuracy to evaluate all models. The ZSL performance is obtained by considering only the subset of test samples from the unseen test classes. For the GZSL performance, the models are evaluated on the full test set which includes seen and unseen classes. We then report the performance on the subsets of seen (S) and unseen (U) classes, and also report their harmonic mean (HM).

\mypara{Implementation details.} 
For \modelName, we use $d_{in_a}=128$, $d_{in_v}=4096$, $d_{fhidd}=512$, $d_{dim}=300$ and $d_{out}=64$. Furthermore, \modelName has $L=6$ transformer layers layers for \ucf and \activity, and $L=8$ for \vgg. We set $d_{pos}=64 $, $d_{ff}=128$. For \activity / \ucf / \vgg we use dropout rates $drop_{enc}=0.1/0.3/0.2$, $drop_{prob,pos}=0.2/0.2/0.1$, $drop_{prob}=0.4/0.3/0.5$, $drop_{proj_w}=0.1/0.1/0.1$, and $drop_{proj_o}=0.1/0.1/0.2$. All attention blocks use $H=8$ heads with a dimension of $d_{head}=64$ per head. 
We train all models using the Adam optimizer \cite{kingma2014adam} with running average coefficients $\beta_1=0.9$, $\beta_2=0.999$, and weight decay $0.00001$. We use a batch size of $64$ for all datasets. 
In order to efficiently train on \activity, we randomly trim the features to a maximum sequence length of 60 during training, and we evaluate on features that have a maximum sequence length of 300 and which are centered in the middle of the video. We note, that \modelName{} can be efficiently trained on a single Nvidia 2080-Ti GPU.
All models are trained for $50$ epochs.  We use a base learning rate of $0.00007$ for \ucf and \activity, and $0.00006$ for \vgg. For \ucf and \activity we use a scheduler that reduces the learning rate by a factor of $0.1$ when the HM on the validation set has not improved for $3$ epochs. To eliminate the bias that the ZSL methods have towards seen classes, we used calibrated stacking~\cite{chao2016empirical} on the search space composed of the interval $[0,3]$ with a step size of $0.2$. 

We train all models with a two-stage training protocol~\cite{mercea2022}. In the first stage, we determine the calibrated stacking~\cite{chao2016empirical} and the epoch with the best HM performance on the validation set. In the second stage, using the hyperparameters from the first stage, we re-train the models on the union of the training and validation sets. We evaluate the final models on the test set.

\subsection{Quantitative results}\label{sec:sota}
We compare our proposed \modelName to state-of-the-art audio-visual ZSL frameworks and to audio-visual frameworks that we adapted to the ZSL task. 

\mypara{Audio-visual ZSL baselines.} We compare our \modelName to three audio-visual ZSL frameworks. \textbf{CJME}~\cite{parida2020coordinated} consists of a relatively simple architecture which maps both input modalities to a shared embedding space. The modality-specific embeddings in the shared embedding space are input to an attention predictor module that determines the dominant modality which is used for the output prediction.
\textbf{AVGZSLNet}~\cite{mazumder2021avgzslnet} builds on CJME by adding a shared decoder and introducing additional loss functions to improve the performance. 
AVGZSLNet removes the attention predictor network and replaces it with a simple average between the output from the head of each modality. 
\textbf{AVCA}~\cite{mercea2022} is a recent state-of-the-art method for audio-visual G(ZSL). It uses a simple cross-attention mechanism on the temporally averaged audio and visual input features to combine the information from the two modalities. Our proposed \modelName{} improves upon the closely related AVCA framework by additionally ingesting temporal information in the audio and visual inputs with an enhanced cross-attention mechanism that gathers information across time and modalities.

\mypara{Audio-visual baselines adapted to ZSL.} We adapt two attention-based audio-visual frameworks to the ZSL setting. \textbf{Attention Fusion}~\cite{fayek2020large} is a method for audio-visual classification which is trained to classify unimodal information. It then fuses the unimodal predictions with learnt attention weights. 
The \textbf{Perceiver}~\cite{jaegle2021perceiver} is a scalable multi-modal transformer framework for flexible learning with arbitrary modality information. It uses a latent bottleneck to encode input information by repeatedly attending to the input with transformer-style attention. The Perceiver allows for a comparison to another transformer-based architecture with focus on multi-modality. We adapt the Perceiver to use the same positional encodings and model capacity as \modelName. We use 64 latent tokens and the same number of layers and dimensions as \modelName. Both Attention Fusion and Perceiver use the same input features, input embedding functions $A_{enc}$ and $V_{enc}$, learning rate and loss functions as \modelName. For Attention Fusion, we temporally average the input features after $A_{enc}$ and $V_{enc}$ to deal with non-synchronous modality sequences due to different feature extraction rates.

All baselines, except for the Perceiver, operate on temporally averaged audio and visual features.
This decreases the amount of information contained in the inputs, in particular regarding the dynamics in a video. In contrast to methods that use temporally averaged inputs, \modelName exploits the temporal dimension which boosts the (G)ZSL performance.

\mypara{Results.}
We compare the results obtained with our \modelName to state-of-the-art baselines for audio-visual (G)ZSL and for audio-visual learning in \cref{tab:final_results_supervised}.
\modelName outperforms all previous methods on the \vgg, \ucf, and \activity datasets for both, GZSL performance (HM) and ZSL performance.
For \activity, our proposed model is significantly better than its strongest competitor AVCA, with a HM of 12.20\% compared to 9.92\% and a ZSL performance of 7.96\% compared to 7.58\%. The CJME and AVGZSLNet frameworks are weaker than the AVCA model.
Similar patterns are exhibited for the \vgg and \ucf datasets.
Interestingly, the GZSL performance for \modelName{} is improved by a more significant margin than the ZSL performance compared to AVCA across all three datasets. This shows that using temporal information and allowing our model to attend across time and modalities is especially beneficial for the GZSL task. 

Furthermore, we observe that the audio-visual Attention Fusion framework and the Perceiver give worse results than AVGZSLNet and AVCA on all three datasets. 
In particular, our \modelName yields stronger ZSL and GZSL performances than the Perceiver which also takes temporal audio and visual features as inputs, with a HM of 8.77\% on \vgg for \modelName compared to 4.93\% for the Perceiver.
Attention Fusion and the Perceiver architecture were not designed for the (G)ZSL setting that uses text as side information. Our proposed training objective, used to also train the Perceiver, aims to regress textual embeddings which might be challenging for the Perceiver given its tight latent bottlenecks.

\begin{table*}[t]
\centering
\setlength{\tabcolsep}{4pt}
\renewcommand{\arraystretch}{1.2}
 \resizebox{\linewidth}{!}{
 \begin{tabular}{l|cccc|cccc|cccc}
 \hline
 Model & \multicolumn{4}{c}{\vgg} & \multicolumn{4}{c}{\ucf} & \multicolumn{4}{c}{\activity} \\
   & S & U & HM & ZSL & S & U & HM & ZSL & S & U & HM & ZSL \\ 
 \hline
 Attention Fusion &14.13  & 3.00 &4.95 &3.37 &39.34 &18.29 &24.97 &20.21 & 11.15 & 3.37 & 5.18&4.88   \\
 Perceiver & 13.25&3.03  &4.93  &3.44 & 46.85& 26.82&34.11  &28.12 & 18.25 &4.27 &6.92  &4.47  \\ \hline
 CJME & 10.86 & 2.22 &3.68 &3.72 &33.89 &24.82 &28.65 &29.01 &10.75 &5.55 &7.32 &6.29  \\
 AVGZSLNet & \textbf{15.02}& 3.19& 5.26& 4.81&\textbf{74.79} &24.15 &36.51 &31.51 &13.70 &5.96 &8.30 &6.39 \\
  AVCA &12.63  &6.19  &8.31 &6.91 & 63.15&30.72 &41.34 &37.72 &16.77  & 7.04 &9.92  &7.58    \\
 \hline
  \modelName & 12.63& \textbf{6.72} &\textbf{8.77} &\textbf{7.41} & 67.14 &\textbf{40.83} &\textbf{50.78} &\textbf{44.64} &\textbf{30.12} &\textbf{7.65} &\textbf{12.20} &\textbf{7.96}  \\
 \hline
 \end{tabular}
 }
 \caption{Performance of our \modelName and of state-of-the-art methods for audio-visual (G)ZSL on the \vgg, \ucf, and \activity datasets. The mean class accuracy for GZSL is reported on the seen (S) and unseen (U) test classes, and their harmonic mean (HM). For the ZSL performance, only the test subset of unseen classes is considered.}
 \label{tab:final_results_supervised}
 \end{table*}

\subsection{Ablation study on the training loss and attention variants}\label{sec:ablation}
Here, we analyse different components of our proposed \modelName. We first compare the performance of our model when trained using different loss functions. We then investigate the influence of the attention mechanisms used in the model architecture on the (G)ZSL performance. Finally, we show that using multi-modal inputs is beneficial and results in outperforming unimodal baselines.

\mypara{Comparing different training losses.}
We show the contributions of the different components in our training loss function to the (G)ZSL performance in \cref{tab:ablation_loss}. 
Using only the regression loss $l_{reg}$ to train our model results in the weakest performance across all datasets, with HM/ZSL performances of 16.25\%/30.17\% on \ucf compared to 50.78\%/44.64\% for our full \modelName. Interestingly, the seen performance (S) when using only $l_{reg}$ is relatively weak, likely caused by the calibrated stacking. Similarly, on \activity, using only $l_{reg}$ yields a low test performance of 0.43\% HM.
Jointly training with the regression and cross-entropy loss functions ($l_{reg}+l_{ce}$) improves the GZSL and ZSL performance significantly, giving a ZSL performance of 4.31\% compared to 2.50\% for $l_{reg}$ on \vgg. The best results are obtained when training with our full training objective $l$ which includes a reconstruction loss term, giving the best performance on all three datasets. 
\begin{table*}[t]
\centering
\setlength{\tabcolsep}{4pt}
\renewcommand{\arraystretch}{1.2}
 \resizebox{\linewidth}{!}{
 \begin{tabular}{l|cccc|cccc|cccc}
 \hline
  Loss & \multicolumn{4}{c}{\vgg} & \multicolumn{4}{c}{\ucf} & \multicolumn{4}{c}{\activity} \\
   & S & U & HM & ZSL & S & U & HM & ZSL & S & U & HM & ZSL \\ \hline
 $l_{reg}$ &0.10  &2.41& 0.19& 2.50& 14.30 &18.82 &16.25  &30.17  &  1.09 & 0.27  & 0.43 & 2.11      \\
 $l_{reg}+l_{ce}$ &\textbf{13.67} &4.06 &6.26 &4.31 &\textbf{75.31}&37.15&49.76&41.75& 11.36& 5.28&7.21 &5.31 \\
 $l=l_{reg}+l_{ce}+l_{rec}$ &12.63 &\textbf{6.72}&\textbf{8.77}&\textbf{7.41}&67.14&\textbf{40.83}&\textbf{50.78} &\textbf{44.64}&\textbf{30.12}&\textbf{7.65}&\textbf{12.20}&\textbf{7.96}\\
 \hline
 \end{tabular}
 }
 \caption{Influence of using different components of our proposed training objective for training \modelName on the (G)ZSL performance on the \vgg, \ucf, and \activity datasets.}
 \label{tab:ablation_loss}
 \end{table*}

\mypara{Comparing different attention variants.}
We study the use of different attention patterns in \cref{tab:ablation_attention}. In particular, we analyse the effect of using within-modality ($\mathbf{A}_{self}$) and cross-modal ($\mathbf{A}_{x}$) attention (cf.\ \cref{eq:attention}), on the GZSL and ZSL performance. Additionally, we investigate models that use a classification token $x^c$ with corresponding output token $c_o$ (\textit{with class. token}) and models for which we simply average the output of the transformer layers which is then used as input to $O_{proj}$ (\textit{w/o class. token}).

Interestingly, we observe that with no global token, using the full attention $\mathbf{A}_{self} + \mathbf{A}_{x}$ gives better results than using only cross-attention on \ucf and \activity for ZSL and GZSL, but is slightly worse on \vgg. This suggests that the bottleneck introduced by limiting the information flow in the attention when using only cross-attention is beneficial for (G)ZSL on \vgg.
When not using the classification token and only self-attention $\mathbf{A}_{self}$, representations inside the transformer are created solely within their respective modalities. 

Using a classification token (\textit{with class. token}) and the cross-attention variant ($\mathbf{A}_c + \mathbf{A}_{x}$) yields the strongest ZSL and GZSL results across all three datasets. The most drastic improvements over full attention can be observed on the \ucf dataset, with a HM of 50.78\% for the cross-attention with classification token ($\mathbf{A}_c + \mathbf{A}_{x}$) compared to 39.18\% for the full attention ($\mathbf{A}_c + \mathbf{A}_{self} + \mathbf{A}_{x}$).
Furthermore, when using $x_c$, cross-attention $\mathbf{A}_{x}$ instead of self-attention $\mathbf{A}_{self}$ leads to a better performance on all three datasets. For $\mathbf{A}_x$ and $x_c$, we obtain HM scores of 8.77\% and 50.78 \% on \vgg and \ucf compared to 6.71\% and 37.37\% with $\mathbf{A}_{self}$ and $x_c$. This shows that using information from both modalities is important for creating strong and transferable video representations for (G)ZSL.
Using the global token relaxes the pure cross-attention setting to a certain extent, since $\mathbf{A}_c$ allows for attention between all tokens from both modalities and the global token. The results in \cref{tab:ablation_attention} have demonstrated the clear benefits of our cross-attention variant used in \modelName.

\begin{table*}[t]
\centering
\setlength{\tabcolsep}{4pt}
\renewcommand{\arraystretch}{1.2}
\resizebox{\linewidth}{!}{
 \begin{tabular}{l|cccc|cccc|cccc}
 \hline
 Model & \multicolumn{4}{c}{\vgg} & \multicolumn{4}{c}{\ucf} & \multicolumn{4}{c}{\activity} \\
  & S & U & HM & ZSL & S & U & HM & ZSL & S & U & HM & ZSL \\\hline
\textit{w/o class. token} & & & & & & & & & & & & \\
$\mathbf{A}_{self} + \mathbf{A}_{x}$ &\textbf{18.40} &3.78 &6.27 &4.25 &31.70 &32.57 &32.13 &33.26 &11.87 &3.80 &5.75 &3.90 \\
$\mathbf{A}_{self}$ &16.08 & 3.56& 5.83& 4.00&42.59 & 24.04&30.73 &27.49 & 9.51 &4.33 &5.95 &4.39 \\
$\mathbf{A}_{x}$ &14.62 &4.22 &6.55 &4.59 & 19.52& 29.80&23.62 &31.35 &1.85 &3.50 &2.42 &3.50 \\\hline
\textit{with class. token} & & & & & & & & & & & & \\
$\mathbf{A}_{c} + \mathbf{A}_{self} + \mathbf{A}_{x}$ &11.36 & 5.50&7.41 &5.97 & 36.73&\textbf{41.99} &39.18 &42.56 & 17.75& 6.79& 9.83&6.89 \\
$\mathbf{A}_{c} + \mathbf{A}_{self}$ &12.23 &4.63 & 6.71&5.25 & 40.14&34.95  &37.37 &35.74  & 4.24 & 3.23 & 3.67 & 3.25 \\
$\mathbf{A}_{c} + \mathbf{A}_{x}$ (\modelName) &12.63 &\textbf{6.72} &\textbf{8.77} &\textbf{7.41} & \textbf{67.14}&40.83 &\textbf{50.78} &\textbf{44.64} &\textbf{30.12} &\textbf{7.65} &\textbf{12.20} &\textbf{7.96}  \\\hline
 \end{tabular}
}
\caption{Ablation of different attention variants with and without a classification token on the \vgg, \ucf, and \activity datasets.}
\label{tab:ablation_attention}
\end{table*}

\mypara{The influence of multi-modality.}
\begin{table*}[t]
\centering
\setlength{\tabcolsep}{4pt}
\renewcommand{\arraystretch}{1.2}
 \resizebox{\linewidth}{!}{
 \begin{tabular}{l|cccc|cccc|cccc}
 \hline
 Model & \multicolumn{4}{c}{\vgg} & \multicolumn{4}{c}{\ucf} & \multicolumn{4}{c}{\activity} \\
  & S & U & HM & ZSL & S & U & HM & ZSL & S & U & HM & ZSL \\\hline
\modelName - audio &5.11 &4.06 &4.53 &4.28 & 35.51& 19.75&25.38 &24.24 & 9.28&4.26& 5.84& 4.65\\
\modelName - visual &3.97 & 3.12 &3.50& 3.19& 38.10&26.84 &31.49 &27.25 & 2.75 & 3.11 & 2.92 &3.11 \\\hline
\modelName &\textbf{12.63} & \textbf{6.72} &\textbf{8.77}&\textbf{7.41} & \textbf{67.14}&\textbf{40.83} &\textbf{50.78} &\textbf{44.64} &\textbf{30.12}& \textbf{7.65} &\textbf{12.20} &\textbf{7.96}  \\
 \hline
 \end{tabular}
 }
 \caption{Influence of using multiple modalities for training and evaluating our proposed model on the (G)ZSL performance on the \vgg, \ucf, and \activity datasets.}
 \label{tab:ablation_unimodal}
 \end{table*}
We compare using only a single input modality for training \modelName to using multiple input modalities in \cref{tab:ablation_unimodal}. For the unimodal baselines \modelName - audio and \modelName - visual, we train \modelName only with the corresponding input modality. 
Using only audio inputs gives stronger GZSL and ZSL results than using only visual inputs on \vgg and \activity. We obtain a HM of 5.84\% for audio compared to 2.92\% for visual inputs on \activity. Interestingly this pattern is reversed for the \ucf dataset where using visual inputs only results in a slightly higher performance than using the audio inputs with HM scores of 31.49\% compared to 25.38\%, and ZSL scores of 27.25\% and 24.24\%.
However, using both modalities (\modelName) increases the HM to 50.78\% and ZSL to 44.64\% on \ucf. Similar trends can be observed for \vgg and \activity which highlights the importance of the tight multi-modal coupling in our \modelName.

\subsection{Qualitative results}\label{sec:qualitative}
\begin{figure}[t]
     \centering
         \includegraphics[width=0.94\textwidth,trim=40 0 2 0, clip]{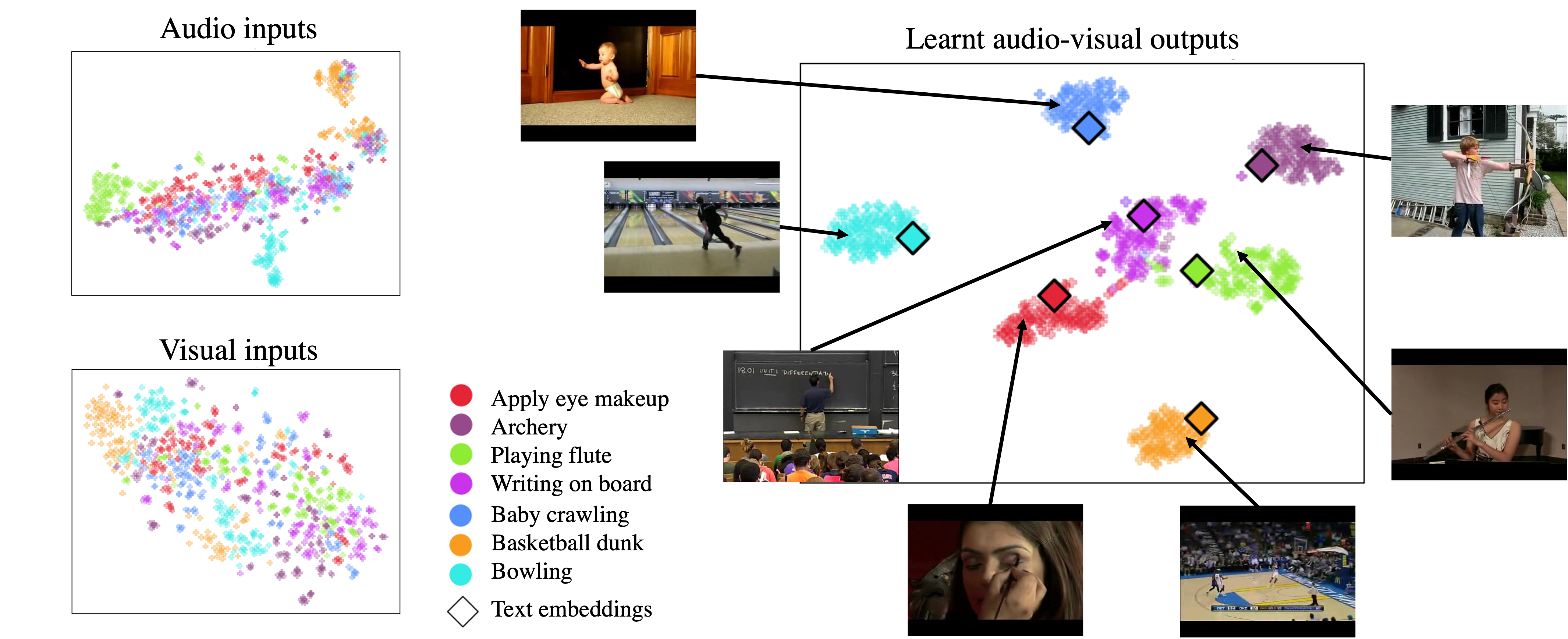}
        \caption{t-SNE visualisation for five seen (\textit{apply eye makeup, archery, baby crawling, basketball dunk, bowling}) and two unseen (\textit{playing flute, writing on board}) test classes from the \ucf dataset, showing audio and visual input embeddings extracted with C3D and VGGish, and audio-visual output embeddings learned with \modelName. Textual class label embeddings are visualised with a square.
        }
        \label{fig:qualitative_results_ucf}
\end{figure}

We present a qualitative analysis of the learnt audio-visual embeddings in \cref{fig:qualitative_results_ucf}. For this, we show t-SNE~\cite{van2008visualizing} visualisations for the audio and visual input features and for the learnt multi-modal embeddings from 7 classes in the \ucf test set. We averaged the input features for both modalities across time. We observe that the audio and visual input features are poorly clustered.
In contrast, the audio-visual embeddings ($\theta_o$) are clearly clustered for both, seen and unseen classes. This suggests that our network is actually learning useful representations for unseen classes, too.
Furthermore, the word2vec class label embeddings ($\theta_w^j$) lie inside the corresponding audio-visual clusters. This confirms that the learnt audio-visual embeddings are mapped to locations that are close to the corresponding word2vec embeddings, showing that our embeddings capture semantic information from the word2vec representations.

\section{Conclusion}
We presented a cross-attention transformer framework that addresses (G)ZSL for video classification using audio-visual input data with temporal information. Our proposed model achieves state-of-the-art performance on the three audio-visual (G)ZSL datasets \ucf, \vgg, and \activity. The use of pre-extracted audio and visual features as inputs results in a computationally efficient framework compared to using raw data. We demonstrated that using cross-modal attention on temporal audio and visual input features and suppressing the contributions from the within-modality self-attention is beneficial for obtaining strong audio-visual embeddings that can transfer information from classes seen during training to novel, unseen classes at test time. 

\mypara{Acknowledgements:} This work was supported by BMBF FKZ: 01IS18039A, DFG: SFB 1233 TP 17 - project number 276693517, by the ERC (853489 - DEXIM), and by EXC number 2064/1 – project number 390727645. The authors thank the International Max Planck Research School for Intelligent Systems (IMPRS-IS) for supporting O.-B. Mercea and T. Hummel. The authors would like to thank M. Mancini for valuable feedback.

%
%


\hyphenation{VGGish}

\makeatletter

\makeatother


\pagestyle{headings}
\mainmatter
\def\ECCVSubNumber{6942}  

\title{Supplementary material: Temporal and cross-modal attention for audio-visual zero-shot learning} 

\titlerunning{Supp. material: Temporal and cross-modal attention for audio-visual ZSL}

\authorrunning{O.-B. Mercea et al.}
%


%
\author{}
%
%
\institute{}
\maketitle

\title{Supplementary material: Temporal and cross-modal attention for audio-visual zero-shot learning} 

In the supplementary material, we provide additional details about baselines (\cref{section:baselines_details}), and present further model ablations (\cref{section:supp_ablations}).  Additionally, we study t-SNE visualisations for \modelName and \cite{mercea2022} (\cref{section:tsne}), and provide a comparison of the computational complexity of \modelName and some of the baselines (\cref{section:computational_complexity}). Finally, we present further quantitative results for audio-visual (G)ZSL when using SeLaVi~\cite{asano2020labelling} features as inputs (\cref{section:selavi_results}).

\section{Additional details about baselines}
\label{section:baselines_details}
In the following, we detail our adaptations of Attention Fusion~\cite{fayek2020large} and of the Perceiver~\cite{jaegle2021perceiver} to the (G)ZSL setting (which we briefly summarised in Section~4.2 of our manuscript).

\subsection{Attention Fusion}
In order to use Attention Fusion~\cite{fayek2020large} in the (G)ZSL setting, we take the same temporal audio and visual features as inputs as \modelName. Following \modelName, we embed the input features into the same feature dimension using $A_{enc}$ and $V_{enc}$. Instead of directly mapping to the number of classes, as the authors originally proposed, $A_{enc}$ and $V_{enc}$ map the features to $\mathbb{R}^{d_{dim}}$. The embedded features are then temporally averaged to obtain a single $d_{dim}-$dimensional feature vector for each modality. The attention weight $\alpha$, which is used for fusing both modalities, is computed using the channel-wise concatenation of the audio and visual embeddings through a linear layer $f_{attn}:\mathbb{R}^{2*d_{dim}} \rightarrow \mathbb{R}^{d_{dim}}$, followed by a sigmoid function. Both modalities are then fused to create the output token $o_c$ through $o_c = \alpha \odot \phi_{a,avg} + (1-\alpha) \odot \phi_{v,avg}$, where $\phi_{a,avg}$ and $\phi_{v,avg}$ are the temporally averaged audio and visual features. $o_c$ is then projected using the same projection function $O_{proj}$, decoder $D_o$, and text embedding projections as in \modelName. We train Attention Fusion using the same learning rate and loss functions as \modelName.
 
\subsection{Perceiver}
The Perceiver~\cite{jaegle2021perceiver} takes the same audio and visual features as input as \modelName. For consistency between frameworks, we again embedded the input features to the same feature dimension using $A_{enc}$ and $V_{enc}$, and equip both \modelName and the Perceiver with the same temporal and modality information by adding positional embeddings as described in the main paper. Our goal was to directly compare our cross-attention mechanism with the Perceiver attention.
Therefore, we adapted the cross-attention, self-attention and dense layer blocks of the Perceiver to use the same internal dimensions as \modelName. We also added a dropout layer at the end of dense layer blocks to match the dense blocks in \modelName. For the randomly initialised latent array, we use $64$ latent tokens with dimension $\mathbb{R}^{d_{dim}}$ for all datasets. Increasing the number of latent tokens did not provide a boost in performance, but significantly increased the computational costs. One of the latent tokens is used as the output classification token $c_o$. We use one cross-attention block and one self-attention block per layer without weight sharing and use the same number of layers as \modelName. This results in just a slightly higher number of parameters for the Perceiver than for our \modelName. The output token $c_o$ is  projected using the projection function $O_{proj}$ and the decoder $D_o$. The computations for the text embeddings are analogous to \modelName. We train the Perceiver using the same learning rate and loss functions as our model.

\section{Additional model ablations}\label{section:supp_ablations}

In this section, we first study the impact of using temporal embeddings (\cref{section:embeddings}) and of the number and design of the cross-attention layers in \modelName (\cref{section:layer_ablation}).
Next, we evaluate the impact on performance when adding noise to the audio modality (\cref{section:noise_impact}).
Finally, we present results of transforming \modelName to \cite{mercea2022} (\cref{section:tcaf_to_avca}).

\subsection{Influence of using temporal information}
\label{section:embeddings}
In the following, we investigate the influence of using temporal information when learning multi-modal video representation for (G)ZSL with \modelName. Since the operations in our audio-visual transformer layers (cf.\ Section~3.2 in the manuscript) are invariant to permutation, the feature tokens are additionally equipped with temporal information through the addition of positional embeddings $pos_t$. Without temporal embeddings, the model is unable to put data from one time step in temporal relation to information from the other time steps. Temporal embeddings therefore allow the model to understand the concept of time. 

\cref{tab:ablation_positional-embedding} shows results for training and evaluating \modelName with ($+$) and without ($-$) temporal embeddings ($pos_t$). The highest harmonic mean is achieved when using temporal embeddings. For instance for \activity, our model that does not use temporal embeddings  ($-pos_t$) obtains only a HM of 8.69\% and a ZSL score of 5.53\%, compared to a HM of 12.20\% and a ZSL score of 7.96\% when using temporal embeddings. Similar observations can be made for \vgg and \ucf, showing the importance of temporal information for learning strong video representations. 

\begin{table*}[t]
\centering
\setlength{\tabcolsep}{4pt}
\renewcommand{\arraystretch}{1.2}
 \resizebox{\linewidth}{!}{
 \begin{tabular}{l|cccc|cccc|cccc}
 \hline
 Positional embeddings & \multicolumn{4}{c}{\vgg} & \multicolumn{4}{c}{\ucf} & \multicolumn{4}{c}{\activity} \\
  & S & U & HM & ZSL & S & U & HM & ZSL & S & U & HM & ZSL \\\hline
$-pos_t$ & \textbf{15.78}&4.66 &7.19 &4.97 &27.35 & 26.02&26.67 &28.06 &21.80 &5.43 &8.69 &5.53 \\
$+pos_t$ (\modelName) &12.63 &\textbf{6.72} &\textbf{8.77} &\textbf{7.41} & \textbf{67.14}&\textbf{40.83} &\textbf{50.78} &\textbf{44.64} &\textbf{30.12} &\textbf{7.65}  &\textbf{12.20} &\textbf{7.96} \\
 \hline
 \end{tabular}
 }
 \caption{Influence of temporal information provided through positional embeddings ($pos_t$) on the (G)ZSL performance on the \vgg, \ucf, and \activity datasets.}
 \label{tab:ablation_positional-embedding}
 \end{table*}

\subsection{Impact of using different amounts of cross-attention layers and of varying the cross-attention layer design}
\label{section:layer_ablation}
In \cref{tab:ablation_layer_number}, we present ablations on the number of cross-attention layers used in our model. Furthermore, we investigate the relevance of using feed forward functions (FF) in our cross-attention layers.

For \modelName, we used 8 cross-attention layers on \vgg (all layers). On the \ucf and \activity datasets, we used 6 layers (all layers). We observe that using more layers is beneficial for GZSL and ZSL performance across all datasets. Moreover, we observe that, in general, eliminating the feed forward functions leads to a decrease in performance. 
Finally, using only half of the layers jointly with self-attention ($1/2*$(all layers) $+A_{self}$) leads to worse overall HM performance than using half of the layers without self-attention ($1/2*$(all layers)). This is in line with the experiments in the main paper, where adding the self-attention leads to worse results. 

This ablation shows that using only cross-attention is beneficial even when using a different number of layers. Furthermore, using more cross-attention layers that are equipped with feed forward functions brings a boost in performance.

\begin{table*}[t]
\centering
\setlength{\tabcolsep}{4pt}
\renewcommand{\arraystretch}{1.2}
 \resizebox{\linewidth}{!}{
 \begin{tabular}{l|cccc|cccc|cccc}
 \hline
 Layer configurations & \multicolumn{4}{c}{\vgg} & \multicolumn{4}{c}{\ucf} & \multicolumn{4}{c}{\activity} \\
  & S & U & HM & ZSL & S & U & HM & ZSL & S & U & HM & ZSL \\\hline
1 layer w/o FF &19.70 &4.47 &7.29 &4.66 &63.30&26.45&37.31&27.85&15.10 &4.59 &7.04 &4.63 \\
1 layer &\textbf{17.95} &4.78 &7.55 &5.13 &40.07&29.40&33.92&29.74& 28.22&4.85 &8.27 &4.89 \\
$1/2*$(all layers) w/o FF & 11.33&4.25 &6.18 &4.59 & 38.72&23.17 &28.99&23.28 & 8.13& 3.35& 4.75&3.40 \\
$1/2*$(all layers) & 12.08&4.69 &6.75 &5.12 &\textbf{77.19} &30.18 &43.40&34.18& 28.65&6.04 &9.98 &6.25 \\
$1/2*$(all layers) $+A_{self}$ &14.62&4.56&6.96&4.97&53.05 & 34.83&42.05&35.84& \textbf{31.38}&5.93 &9.97 &6.51 \\
all layers w/o FF &14.41 &4.28 &6.60 & 4.59 &32.57&25.77&28.78&28.86& 7.44& 3.27&4.54 &3.33 \\
all layers &12.63 &\textbf{6.72} &\textbf{8.77} &\textbf{7.41}  &67.14 &\textbf{40.83} &\textbf{50.78}& \textbf{44.64}&30.12 &\textbf{7.65} &\textbf{12.20} &\textbf{7.96} \\
 \hline
 \end{tabular}
 }
 \caption{Varying the number of cross-attention layers in \modelName and the use of feed forward (FF) functions in the cross-attention layers.}
 \label{tab:ablation_layer_number}
 \end{table*}

\subsection{Impact of noise in audio stream on GZSL performance}
\label{section:noise_impact}
\begin{figure}{}{}
  \begin{center}
     \includegraphics[width=1.0\textwidth]{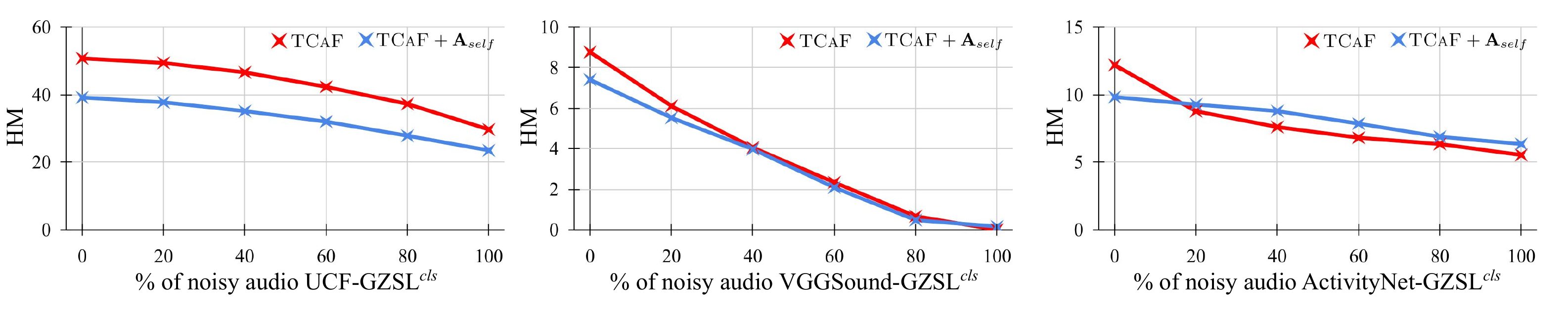}
  \end{center}
  \caption{Robustness of \modelName and \modelName+$A_{self}$ to noise added to different proportions of the audio stream on \ucf, \vgg and \activity.}\label{fig:supp_audiowithnoise_ucf}
\end{figure}

In this section, we study how the GZSL performance (HM) of \modelName decreases when noise is added to increasing temporal portions of the audio signal on all three datasets. We study both \modelName and \modelName+$A_{self}$ in \cref{fig:supp_audiowithnoise_ucf}. It can be observed that an increase in the proportion of noise leads to a decrease in the GZSL performance for both models on all three datasets. Furthermore, it can be observed that \modelName is significantly more robust to perturbations on \ucf and slightly more robust on \vgg. On the other hand, we can observe that on \activity the trend is reversed, with \modelName+$A_{self}$ being slightly more robust. Overall, it can be argued that \modelName is more robust across all three datasets than \modelName+$A_{self}$.

\subsection{Transforming \modelName into \cite{mercea2022}}
\label{section:tcaf_to_avca}
Our \modelName builds on the AVCA~\cite{mercea2022} framework for audio-visual GZSL. To highlight the benefits of \modelName compared to AVCA, we show results for transforming \modelName into AVCA~\cite{mercea2022} in \cref{tab:ablation_avca_tcaf}. 

\modelName exploits temporal information and obtains a HM performance of 8.77\% on \vgg compared to a HM of 7.65\% (\modelName avg input) when using temporally averaged inputs. Moreover, \modelName uses an enhanced cross-modal attention to effectively gather multi-modal information. On the other hand, the attention mechanism of \cite{mercea2022} uses temporally averaged feature inputs, which leads to a HM of 6.82\% on \vgg (\cite{mercea2022}). Additionally, \modelName uses a single output branch and a classification token to aggregate the multi-modal information. In contrast, \cite{mercea2022} uses two branches and no classification token which leads to a HM of 6.27\% (w/o class. token) on \vgg. Finally, our training objective avoids triplet losses, i.e. there is no overhead to train with positive and negative pairs. Using triplet losses similar to those used in \cite{mercea2022} leads to a lower performance (\modelName + $l_{triplet}$) than \modelName.
The same trend can be observed for the other datasets, proving that our architectural choices are more suitable for the audio-visual (G)ZSL task.

\begin{table*}[t]
\centering
\setlength{\tabcolsep}{4pt}
\renewcommand{\arraystretch}{1.2}
 \resizebox{\linewidth}{!}{
 \begin{tabular}{l|cccc|cccc|cccc}
 \hline
 Model & \multicolumn{4}{c}{\vgg} & \multicolumn{4}{c}{\ucf} & \multicolumn{4}{c}{\activity} \\
  & S & U & HM & ZSL & S & U & HM & ZSL & S & U & HM & ZSL \\\hline
\cite{mercea2022} &12.63&6.19&8.31&6.91&63.15&30.72&41.34&37.72&16.77&7.04&9.92&7.58 \\
\modelName+att from \cite{mercea2022} &10.08&5.16&6.82&5.41&39.47&28.85&33.33&29.79&5.58&2.37&3.33&2.43 \\
\modelName avg input &11.69&5.69&7.65&6.16&12.00&20.46&15.13&20.59&16.43&3.26&5.44&3.42 \\
w/o class. token&\textbf{18.40}&3.78&6.27&4.25&31.70&32.57&32.13&33.26&11.87&3.80&5.75&3.90 \\
\modelName+$l_{triplet}$ &14.51&4.78&7.19&5.06&\textbf{71.61}&35.91&47.83&40.00&18.74&6.58&9.74&6.63 \\
\modelName &12.63&\textbf{6.72}&\textbf{8.77}&\textbf{7.41}&67.14&\textbf{40.83}&\textbf{50.78}&\textbf{44.64}&\textbf{30.12}&\textbf{7.65}&\textbf{12.20}&\textbf{7.96}\\
 \hline
 \end{tabular}
 }
 \caption{Transforming \modelName into \cite{mercea2022}}
 \label{tab:ablation_avca_tcaf}
 \end{table*}

\section{t-SNE comparison between \modelName and \cite{mercea2022}}
\label{section:tsne}
\begin{figure}[t]
     \centering
         \includegraphics[width=0.94\textwidth,trim=5 0 2 0, clip]{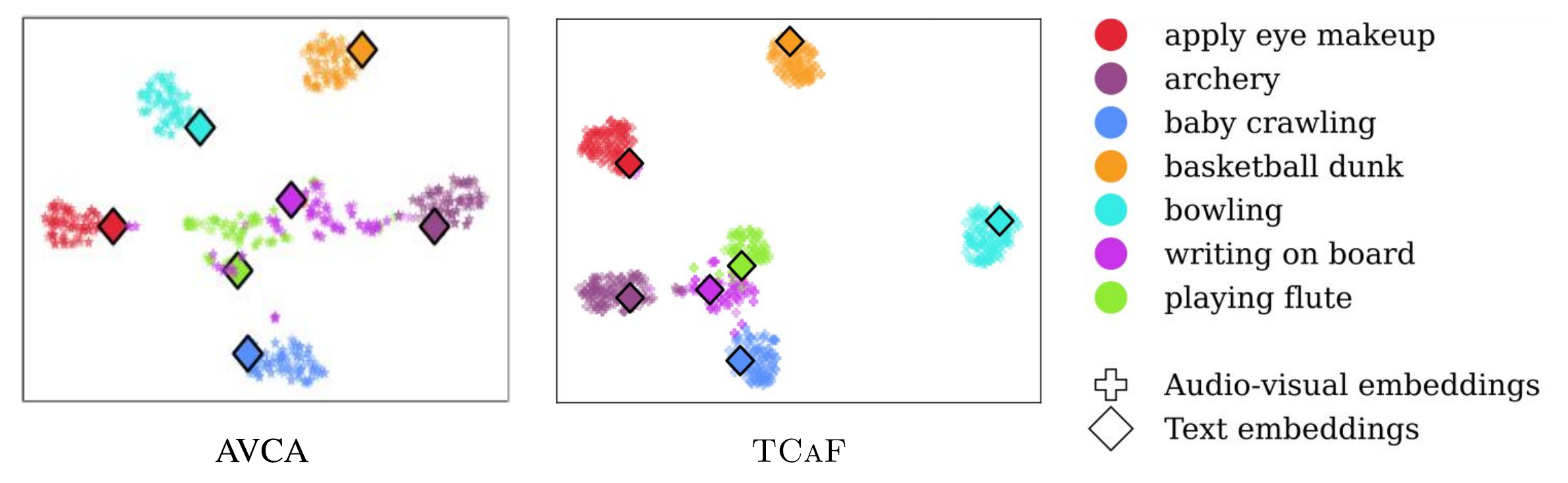}
        \caption{t-SNE visualisations for five seen (\textit{apply eye makeup, archery, baby crawling, basketball dunk, bowling}) and two unseen (\textit{playing flute, writing on board}) test classes from the \selaucf dataset, showing the difference between \modelName and \cite{mercea2022}. Textual class label embeddings are visualised with a square.
        }
        \label{fig:qualitative_results_tcaf_avca}
\end{figure}

We show t-SNE visualisations that highlight the difference between \modelName and \cite{mercea2022} in \cref{fig:qualitative_results_tcaf_avca}. It can be observed that in the case of \cite{mercea2022}, the classes overlap more than in the case of \modelName. In particular, this can be observed for the unseen classes. Moreover, for \cite{mercea2022}, the clusters are less concentrated than for \modelName. 

\section{Computational complexity}
\label{section:computational_complexity}
The computational complexity increases with the length
of the temporal sequence. Using the average duration of the data in \ucf and
a single forward pass for a batch of 256 samples, \modelName
requires 51.8 GFLOPS vs 174.1 for~\cite{jaegle2021perceiver} and 4.4 for~\cite{mercea2022}. The Perceiver~\cite{jaegle2021perceiver} uses a transformer architecture along with the temporal dimension, while~\cite{mercea2022} does not use the temporal dimension. Thus, it can be observed that \modelName is more resource-efficient than the most similar baseline.
\modelName was trained on a single NVIDIA 2080Ti GPU.

\section{Additional quantitative results with SeLaVi~\cite{asano2020labelling} features}
\label{section:selavi_results}
In this section, we present additional results that show the performance of our \modelName with SeLaVi~\cite{asano2020labelling} input features from \cite{mercea2022} in Table \ref{tab:final_results_selavi}. On \selavgg, \modelName obtains a HM of 7.33\% compared to 6.31\% for AVCA and a ZSL of 6.06\% for \modelName vs.\ 6.00\% for AVCA. Furthermore, on \selaucf, \modelName significantly outperforms AVCA, with a HM of 31.72\% compared to 27.15\% and a ZSL performance of 24.81\% compared to 20.01\% for AVCA. On the other hand, on \selaactivity, AVCA outperforms \modelName with a HM of 12.13\% vs 10.71\% for \modelName and a ZSL of 9.13\% for AVCA vs 7.91\% for \modelName. However, on \selaactivity, \modelName outperforms Perceiver which is the most similar baseline to \modelName and which also uses temporal features.

\begin{table}[h]
\centering
\setlength{\tabcolsep}{4pt}
\renewcommand{\arraystretch}{1.2}
 \resizebox{\linewidth}{!}{
 \begin{tabular}{l|cccc|cccc|cccc}
 \hline
 Model & \multicolumn{4}{c}{\selavgg} & \multicolumn{4}{c}{\selaucf} & \multicolumn{4}{c}{\selaactivity} \\
   & S & U & HM & ZSL & S & U & HM & ZSL & S & U & HM & ZSL \\ 
 \hline
 Att. Fusion &6.12&2.26&3.30&2.38&35.47&11.26&17.10&12.54&6.49&2.04&3.11&2.63   \\
 Perceiver &7.92&2.72&4.05&2.93&34.10&18.18&23.72&18.77&7.22&5.16&6.02&5.37  \\ \hline
 CJME &8.69&4.78&6.17&5.16&26.04&8.21&12.48&8.29&5.55&4.75&5.12&5.84  \\
 AVGZSLNet &\textbf{18.05}&3.48&5.83&5.28&52.52&10.90&18.05&13.65&8.93&5.04&6.44&5.40 \\
  AVCA &14.90&4.00&6.31&6.00&51.53&18.43&27.15&20.01&\textbf{24.86}&\textbf{8.02}&\textbf{12.13}&\textbf{9.13}    \\
 \hline
  \modelName (ours) &9.64&\textbf{5.91}&\textbf{7.33}&\textbf{6.06}&\textbf{58.60}&\textbf{21.74}&\textbf{31.72}&\textbf{24.81}&18.70&7.50&10.71&7.91  \\
 \hline
 \end{tabular}
 }
 \caption{Audio-visual (G)ZSL results when using SeLaVi~\cite{asano2020labelling} audio and visual features as inputs on the \selaactivity, \selavgg, and \selaucf datasets.}
 \label{tab:final_results_selavi}
 \end{table}

%
%
\bibliographystyle{splncs04}
\bibliography{egbib}
\end{document}